\documentclass{article} 
\usepackage{iclr2027_conference,times}

\usepackage{amsmath,amsfonts,bm}

\def\eqref#1{equation~\ref{#1}}

\def\1{\bm{1}}

\DeclareMathAlphabet{\mathsfit}{\encodingdefault}{\sfdefault}{m}{sl}
\SetMathAlphabet{\mathsfit}{bold}{\encodingdefault}{\sfdefault}{bx}{n}

\usepackage{times}
\usepackage{amsmath,amsfonts,amssymb,bm}
\usepackage{graphicx}
\usepackage{booktabs}
\usepackage{array}
\usepackage{url}
\usepackage[hidelinks]{hyperref}
\usepackage{lineno}
\usepackage{comment}
\usepackage{wrapfig}
\usepackage{multirow,multicol}
\usepackage{enumitem}
\usepackage{xspace} 
\usepackage{subcaption}
\usepackage{cleveref}

\newcommand{\Dfive}{\ensuremath{\mathrm{D@5}}}
\newcommand{\DfiveNormal}{\ensuremath{\mathrm{D@5}^{\mathrm{N}}}}
\newcommand{\DfiveExcluded}{\ensuremath{\mathrm{D@5}^{\mathrm{X}}}}

\newcommand{\DCDisc}{\ensuremath{\mathrm{DC}\text{-}\mathrm{Disc}}}
\newcommand{\DCDetDelta}{\ensuremath{\Delta\mathrm{DC}\text{-}\mathrm{Det}}}
\newcommand{\DCSelDelta}{\ensuremath{\Delta\mathrm{DC}\text{-}\mathrm{Sel}}}

\newcommand{\bDCDisc}{\ensuremath{\mathbf{DC}\text{-}\mathbf{Disc}}}
\newcommand{\bDCDetDelta}{\ensuremath{\boldsymbol{\Delta}\mathbf{DC}\text{-}\mathbf{Det}}}
\newcommand{\bDCSelDelta}{\ensuremath{\boldsymbol{\Delta}\mathbf{DC}\text{-}\mathbf{Sel}}}

\title{Rethinking Open-World Video Anomaly \\
 Detection: Diagnosing Definition Blindness}

\author{
Inpyo Song \\
SungKyunKwan University \\
South Korea \\
\texttt{songinpyo@skku.edu}
\And
Jangwon Lee \\
SungKyunKwan University \\
South Korea \\
\texttt{leejang@skku.edu}
}

\newif\ifarxiv
\arxivtrue          

\ifarxiv
  \iclrfinalcopy    
\fi

\begin{document}

\maketitle

\ifarxiv
  \fancyhead{}      
\fi

\begin{abstract}
Open-world video anomaly detection (OWVAD) aims to localize anomalous events in videos according to a user-specified definition.
An OWVAD detector should therefore use the supplied definition to decide what is anomalous.
We call this behavior \textbf{definition following}, and its failure \textbf{definition blindness}.
However, existing temporal evaluation protocols may obscure this distinction, making strong generic anomaly detection appear to reflect definition following.
For example, Drift@5~\citep{lagovad} and the customizable VAD protocol~\citep{anyanomaly} treat both normal frames and anomalies outside the current definition as negatives.
Across the three benchmarks used in our study, normal frames account for 78.6\%--96.4\% of these negatives.
As a result, the reported score can be driven largely by target-versus-normal separation rather than target-versus-other-anomaly separation.
Consistent with this diagnosis, a LaGoVAD~\citep{lagovad} branch that never receives the anomaly definition nearly matches the full model under Drift@5.
We therefore introduce three complementary definition-conditioned evaluations
that separate definition following from generic anomaly detection.
Applying them to VAD systems and general vision--language models on UCF-Crime, XD-Violence, and MSAD reveals that strong performance under existing evaluation can coexist with weak definition following.
Motivated by these results, we introduce Definition-Contrastive Scoring (DeCoS), 
which reduces anomaly evidence shared across competing definitions and improves definition following across the three benchmarks.
Our results show that generic anomaly detection and definition following are distinct capabilities and should be evaluated separately in OWVAD.
\end{abstract}

\section{Introduction}
\label{sec:intro}
\begin{figure}[t]
  \centering
  \includegraphics[width=\linewidth]{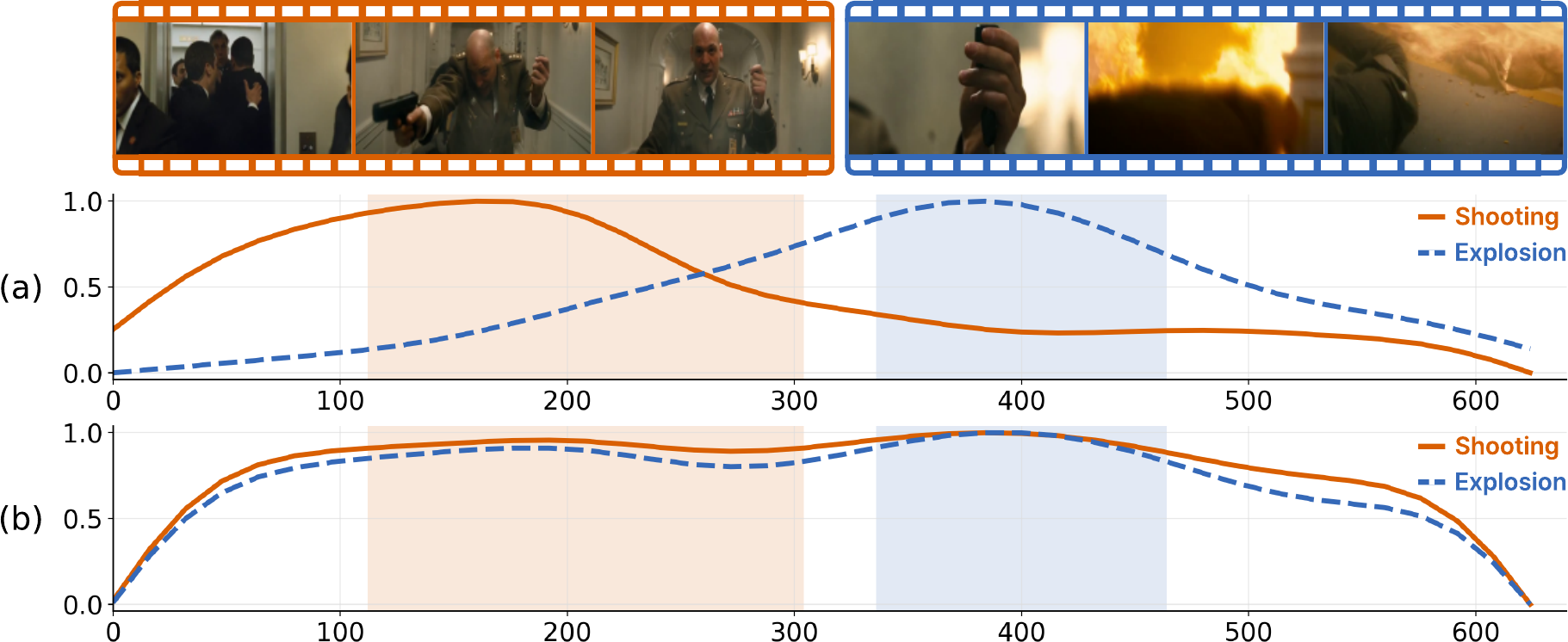}
  \caption{
    \textbf{Definition following and blindness in OWVAD.}
    (a) DeCoS (Ours) assigns higher frame-level scores
    to the event specified by the definition.
    (b) LaGoVAD~\citep{lagovad} produces nearly identical
    score curves for the two definitions.
  }
  \label{fig:dissociation}
\end{figure}

Video anomaly detection (VAD) aims to localize anomalous events in videos~\citep{wu2024vadclip,song2025anomaly}.
In conventional VAD, which events are considered anomalous is predefined in the benchmark~\citep{sultani2018,wu2020not,zhu2024advancing}.
Open-world video anomaly detection (OWVAD), by contrast, allows users to specify which events should be considered anomalous~\citep{lagovad,anyanomaly}.
The detector takes this user-specified anomaly definition as input.
OWVAD therefore requires more than detecting anomalous events.
For the same video, changing the anomaly definition should change which events receive high scores.
Figure~\ref{fig:dissociation} illustrates this distinction with separate shooting and explosion intervals.
In the top row (a), changing the definition from shooting to explosion shifts the highest scores from the shooting interval to the explosion interval.
We call this behavior \textbf{definition following}.
In the bottom row (b), the two definitions produce nearly identical score curves, with both anomalous intervals receiving high scores.
We call this behavior \textbf{definition blindness}.

However, existing VAD evaluation may not clearly distinguish definition following from definition blindness.
Standard VAD metrics operate under a fixed benchmark definition and therefore do not compare anomaly scores across different anomaly definitions.
Recently introduced OWVAD evaluation protocols, such as Drift@5~\citep{lagovad} and customizable VAD (C-VAD) evaluation~\citep{anyanomaly}, specify which anomaly classes are treated as targets.
Yet both compare target anomalies against a negative set that contains normal frames and anomalies excluded by the current definition.
A high score can therefore come from separating target anomalies from normal frames, even if the model does not distinguish target anomalies from other anomalies present in the video.

We find that this imbalance is substantial on current VAD benchmarks.
Across UCF-Crime, XD-Violence, and MSAD, normal frames make up 87.9\%--96.4\% of the negative set under Drift@5 and 78.6\%--93.0\% under class-wise C-VAD evaluation.
To test whether this imbalance can hide definition blindness in practice,
we evaluate a branch of LaGoVAD~\citep{lagovad} that never receives the anomaly definition.
Although this branch produces the same scores for every anomaly definition, it nearly matches the full model under Drift@5, with gaps of only 0.6–1.2 AUROC points and 0.3–0.5 AP points on XD-Violence and MSAD.
These results reveal that even evaluation protocols with varying anomaly definitions may fail to distinguish definition following from definition blindness.
We therefore introduce three complementary definition-conditioned metrics to evaluate definition following: \DCDisc{}, \DCDetDelta{}, and \DCSelDelta{}.
They test whether a model distinguishes the specified anomaly from other anomalies,
detects the specified anomaly better with the matching definition than with non-matching definitions,
and selects the matching event when multiple anomalies occur.
These metrics separate definition following from generic anomaly detection.

We apply these metrics to VAD systems spanning conventional, open-vocabulary, and open-world settings, together with general vision–language models, on UCF-Crime, XD-Violence, and MSAD.
Across our experiments, we find that models that perform well under existing VAD evaluation often struggle to follow the supplied anomaly definition.
Motivated by this gap, we introduce Definition-Contrastive Scoring (DeCoS), which scores the supplied definition relative to competing definitions and reduces anomaly evidence shared across them. DeCoS improves all three definition-conditioned metrics across the three benchmarks.

This study provides new insights into the evaluation of OWVAD by showing that strong generic anomaly detection performance can mask definition blindness, 
even when anomaly definitions vary during evaluation. 
Our core contribution is to make the distinction between anomaly detection and definition following explicit and measurable, 
through dedicated evaluation metrics, empirical analysis, and Definition-Contrastive Scoring. 
Together, these contributions highlight that progress in OWVAD should be measured not only by how well models detect anomalies, but also by whether their predictions follow the supplied anomaly definition.

\section{Related Work}
\label{sec:related_work}

\paragraph{Video anomaly detection.}
Semi-supervised VAD learns from normal training videos~\citep{hasan2016learning,liu2018future,gong2019memorizing,song2026bounding},
while weakly supervised VAD uses video-level labels to localize anomalous events~\citep{sultani2018,rtfm,feng2021mist,song2025anomaly}.
These settings are typically evaluated under a fixed benchmark definition of anomaly~\citep{sultani2018,wu2020not,zhu2024advancing}.
Open-set and open-vocabulary methods extend detection to unseen anomaly classes~\citep{zhu2022openset,ovvad,plovad,anomize},
while cross-domain and multi-domain methods study generalization across datasets and anomaly labels~\citep{crossdomainvad,multidomainvad}.
These lines of work ask whether anomaly detection generalizes to new classes or domains.
We instead study whether temporal anomaly scores follow a supplied definition.

\paragraph{Language-based anomaly detection.}
CLIP-based VAD uses image--text alignment to localize and recognize anomalous events~\citep{radford2021learning,wu2024vadclip,alertclip}.
Multimodal language-model methods use captions and reasoning to detect and explain anomalies~\citep{lavad,vera,holmesvau,song2024video,song2026instance}.
These approaches establish language as a useful source of supervision or conditioning for anomaly detection.
Our method uses language for a different purpose:
it constructs temporal anomaly scores relative to competing definitions and trains the readout to distinguish the matching definition from non-matching definitions.

\paragraph{User-defined anomaly detection and evaluation.}
Recent work extends VAD to user-specified anomaly definitions~\citep{lagovad,anyanomaly,esom}.
In this setting, Drift@5~\citep{lagovad} and the C-VAD protocol~\citep{anyanomaly} 
assess model performance as the user-provided anomaly definition changes.
These protocols, however, do not isolate how the supplied definition changes the ranking of anomaly scores, 
because their evaluations reward generic anomaly detection more heavily than definition following.
Our work separates these effects and evaluates definition following directly.

\section{Diagnosing Definition Blindness}
\label{sec:diagnosing_definition_blindness}

Given a video \(V\) and an anomaly definition \(z\),
an OWVAD detector assigns an anomaly score to each frame:
\begin{equation}
  \label{eq:anomaly_scores}
  S_\tau(V,z) \in \mathbb R,
  \qquad \tau=1,\ldots,|V|,
\end{equation}
where \(|V|\) is the number of frames in \(V\).
Following \citet{lagovad}, 
we assume that abnormality is determined
by the video and the supplied definition.
We write the frame-level target as
\begin{equation}
  \label{eq:frame_target}
  y_\tau^z = F_\tau(V,z) \in \{0,1\},
\end{equation}
where \(y_\tau^z=1\) means that the frame at time \(\tau\)
is anomalous under \(z\), and \(y_\tau^z=0\) otherwise.
Definition following requires the detector to assign higher anomaly scores
to frames that are anomalous under the supplied definition
than to frames that are not.

A clear case of definition blindness occurs when
the score function does not depend on the supplied definition:
\begin{equation}
  \label{eq:definition_blind_scores}
  S_\tau(V,z)=a_\tau(V)
  \qquad
  \forall \tau,\ \forall z,
\end{equation}
where \(a_\tau(V)\) depends only on the video.
Such a detector can localize anomalous events,
but cannot adjust which events receive high scores
to match the supplied definition.

\subsection{Why existing evaluation may conceal definition blindness}
\label{sec:drift5_failure}

In this section, we show why definition-blind detectors can perform well
under existing VAD evaluation.
We identify each evaluation frame by $i=(n,\tau)$,
where $n$ indexes videos $V_n$ and $\tau$ indexes a frame within it.
We write $s_i(z)=S_\tau(V_n,z)$
and use $s(z)$ to denote the collection of these scores.
All frame sets below use these indices.
Let $\mathcal A$ and $\mathcal N$ denote the index sets
of frames labeled anomalous and normal by the benchmark, respectively.
Let $m$ denote either the AUROC or AP evaluation function.
We write $m(s(z);\mathcal P^+,\mathcal P^-)$ for its value with
$\mathcal P^+$ as the positive set and $\mathcal P^-$ as the negative set,
using their scores in $s(z)$.

Standard VAD evaluation treats all anomalous frames as positives
($\mathcal P^+=\mathcal A$) and normal frames as negatives
($\mathcal P^-=\mathcal N$).
It measures generic anomaly detection under a fixed benchmark definition.
Consider a detector that reuses the scores obtained under this definition
for every definition, as in \Cref{eq:definition_blind_scores}.
Its standard VAD AUROC and AP remain unchanged despite its definition blindness.
High performance under this evaluation therefore does not establish definition following.

Drift@5~\citep{lagovad} and the C-VAD protocol~\citep{anyanomaly} instead vary 
which anomaly classes are treated as targets.
Drift@5 selects subsets of classes,
whereas C-VAD evaluates one class at a time.
Let $\mathcal C$ denote the benchmark's anomaly classes.
For each $c\in\mathcal C$, let $\mathcal A_c$ be the set
of indices $i=(n,\tau)\in\mathcal A$
for frames labeled as class $c$.
Under the single-label evaluation protocol considered here,
these sets partition $\mathcal A$.
For a target class set $\Gamma\subseteq\mathcal C$,
let $z_\Gamma$ specify the classes in $\Gamma$ as anomalous.
The target and excluded anomaly frame sets are
\begin{equation}
  \label{eq:definition_partitions}
  \mathcal A_\Gamma=\bigcup_{c\in\Gamma}\mathcal A_c,
  \qquad
  \mathcal A_{\neg\Gamma}=\mathcal A\setminus\mathcal A_\Gamma.
\end{equation}
Under this class-based labeling,
$F_\tau(V_n,z_\Gamma)=1$ exactly when
$i=(n,\tau)\in\mathcal A_\Gamma$ (\Cref{eq:frame_target}).
Both protocols treat $\mathcal A_\Gamma$ as positives
and $\mathcal N\cup\mathcal A_{\neg\Gamma}$ as negatives.
We analyze both AUROC and AP with these labels
and denote the target-versus-rest metric by
\begin{equation}
  \label{eq:target_rest}
  m^{\mathrm{rest}}(\Gamma)
  =
  m\bigl(
    s(z_\Gamma);
    \mathcal A_\Gamma,
    \mathcal N\cup\mathcal A_{\neg\Gamma}
  \bigr).
\end{equation}
This evaluation combines separation of target anomalies from normal frames
with separation from anomalies excluded by the definition.

For AUROC, the overall value is an exact weighted average
of the two comparisons.
Assume that $\mathcal A_\Gamma$, $\mathcal N$,
and $\mathcal A_{\neg\Gamma}$ are nonempty.
Keeping the target frames and scores $s(z_\Gamma)$ unchanged,
we evaluate the two negative sets separately:
\begin{equation}
  \label{eq:negative_components}
  \begin{aligned}
    m^{\mathrm{normal}}(\Gamma)
    &=
    m\bigl(
      s(z_\Gamma);\mathcal A_\Gamma,\mathcal N
    \bigr),\\
    m^{\mathrm{excluded}}(\Gamma)
    &=
    m\bigl(
      s(z_\Gamma);\mathcal A_\Gamma,\mathcal A_{\neg\Gamma}
    \bigr).
  \end{aligned}
\end{equation}
Each target frame with index $i\in\mathcal A_\Gamma$
is compared with every negative frame in
$\mathcal N\cup\mathcal A_{\neg\Gamma}$.
AUROC averages these comparisons,
assigning a value of one when the target score is higher,
one half for a tie, and zero otherwise~\citep{song2026vlm}.
The fraction of comparisons involving a normal negative frame is
\begin{equation}
  \label{eq:normal_negative_fraction}
  \omega_\Gamma
  =
  \frac{|\mathcal N|}
       {|\mathcal N|+|\mathcal A_{\neg\Gamma}|}.
\end{equation}
Grouping the comparisons by the negative frame's set gives
\begin{equation}
  \label{eq:auroc_mixture}
  \begin{aligned}
    \mathrm{AUROC}^{\mathrm{rest}}(\Gamma)
    &=
    \omega_\Gamma 
    \mathrm{AUROC}^{\mathrm{normal}}(\Gamma)
    +
    (1-\omega_\Gamma)
    \mathrm{AUROC}^{\mathrm{excluded}}(\Gamma).
  \end{aligned}
\end{equation}
When $\omega_\Gamma$ is large,
a definition-blind detector can obtain high overall AUROC
by separating anomalies from normal frames,
even if it poorly distinguishes target anomalies from excluded anomalies.

AP has the same negative-pool problem, although it does not decompose into a weighted average as AUROC does.
Normal frames and excluded anomalies both contribute false positives to the same precision calculation.
As a result, lowering normal-frame scores can increase AP even when the scores of all anomalous frames remain unchanged.
High AP under target-versus-rest evaluation therefore does not by itself establish improved definition following.
We provide the exact expression in Appendix~\ref{app:ap_decomposition}.

Normal frames dominate the negative pool under both Drift@5 and C-VAD-style evaluation on current VAD benchmarks.
For Drift@5, normal frames account for $96.4\%$, $87.9\%$, and $88.5\%$ of the negative pool on UCF-Crime, XD-Violence, and MSAD, respectively.
For AUROC, this means that comparisons with excluded anomalies receive only $3.6\%$--$12.1\%$ of the total weight.
Under C-VAD evaluation, the corresponding normal-frame fractions are $93.0\%$, $80.2\%$, and $78.6\%$.
Thus, in both protocols, target-versus-normal discrimination can contribute substantially to the reported score even when the detector does not distinguish the target anomaly from other anomalies.

\begin{wraptable}{r}{0.44\linewidth}
  \vspace{-1.15em}
  \centering
  \scriptsize
  \caption{
  \textbf{A definition-blind branch nearly matches full LaGoVAD.}
  Definition-blind denotes the pre-fusion branch that never receives the queried definition.
}
  \label{tab:lagovad_text_blind}
  \setlength{\tabcolsep}{2.6pt}
  \resizebox{\linewidth}{!}{
    \begin{tabular}{lccccc}
      \toprule
      \multirow{2}{*}{Dataset}
      & \multirow{2}{*}{Score}
      & \multicolumn{2}{c}{Standard VAD}
      & \multicolumn{2}{c}{\Dfive{}}
      \\
      \cmidrule(lr){3-4}
      \cmidrule(lr){5-6}
      &
      & AUROC & AP
      & AUROC & AP
      \\
      \midrule

      \multirow{3}{*}{XD}
      & LaGoVAD
      & 89.5 & 71.5
      & 85.9 & 41.9
      \\
      & Definition-blind
      & 88.1 & 70.6
      & 84.7 & 41.5
      \\
      & $\Delta$
      & \textbf{1.4} & \textbf{0.8}
      & \textbf{1.2} & \textbf{0.5}
      \\

      \midrule

      \multirow{3}{*}{MSAD}
      & LaGoVAD
      & 90.0 & 66.1
      & 85.2 & 37.3
      \\
      & Definition-blind
      & 89.5 & 65.5
      & 84.7 & 37.0
      \\
      & $\Delta$
      & \textbf{0.6} & \textbf{0.5}
      & \textbf{0.6} & \textbf{0.3}
      \\

      \bottomrule
    \end{tabular}
  }
  \vspace{-2em}
\end{wraptable}
We next test whether a definition-blind detector can retain strong performance under Drift@5.
We compare the full LaGoVAD model~\citep{lagovad} with its pre-fusion branch, which never receives an anomaly definition and therefore produces the same frame scores for every definition.
As shown in \Cref{tab:lagovad_text_blind}, this branch nearly matches the full model under both standard VAD evaluation and Drift@5.
On XD-Violence, the Drift@5 gaps are only $1.2$ AUROC points and $0.5$ AP points.
On MSAD, they shrink to $0.6$ AUROC points and $0.3$ AP points.
Thus, strong Drift@5 performance can coexist with complete definition blindness.

\begin{figure*}[t]
  \centering
  \includegraphics[width=0.9\textwidth]{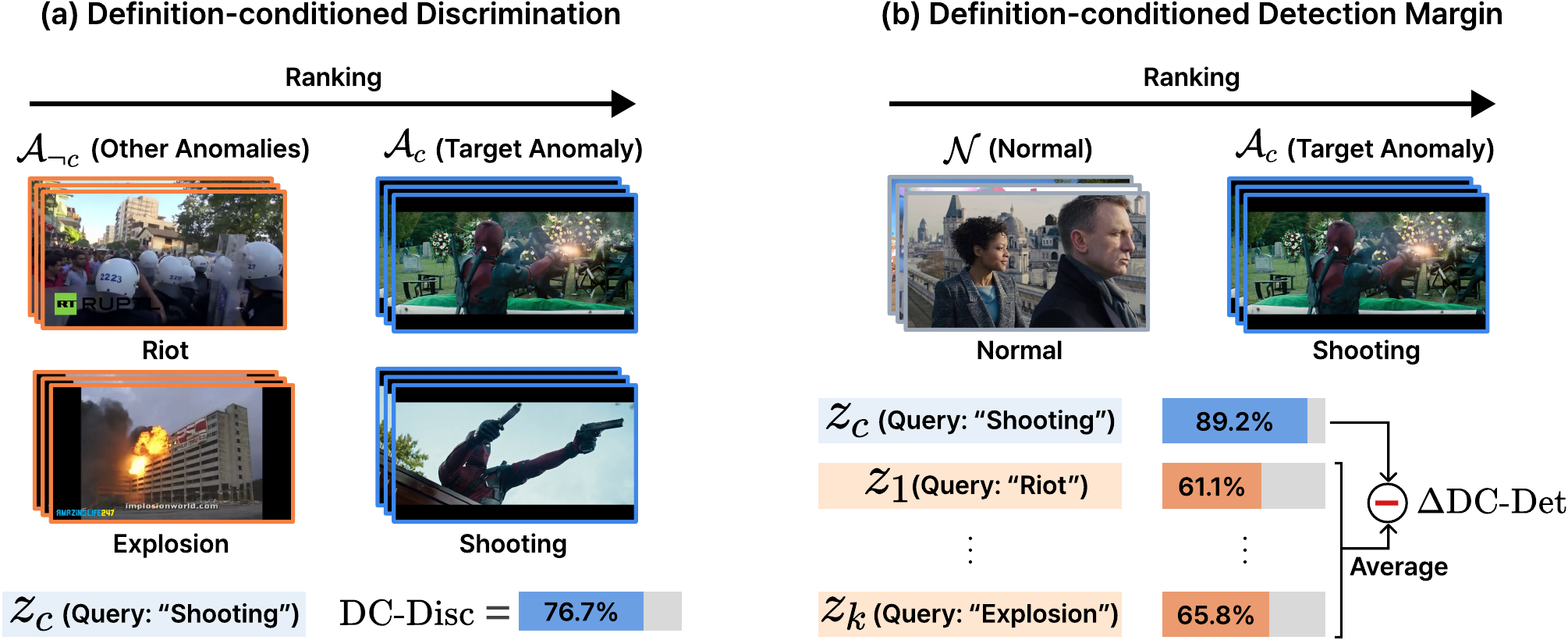}
  \caption{
  \textbf{Illustration of definition-conditioned evaluation for an example target class.}
  (a) \DCDisc{} measures whether the supplied definition ranks target anomaly frames above other anomalies, excluding normal frames.
  and
  (b) \DCDetDelta{} measures the performance difference of the matching definition over the non-matching average on the same anomaly and normal frames.
  }
  \label{fig:metrics}
  \vspace{-1em}
\end{figure*}

\subsection{Definition-conditioned evaluation}
\label{sec:definition_following_metrics}
The analysis above suggests two complementary ways to isolate definition following:
remove normal frames from the comparison,
or hold the evaluated frames fixed and change only the supplied definition.
\DCDisc{} follows the first approach.
\DCDetDelta{} and \DCSelDelta{} follow the second,
measuring the difference from the matching definition for detection and multi-event selection, respectively.
We report all three using base metric $m\in\{\text{AUROC, AP}\}$.

\paragraph{\bDCDisc{}: distinguish target anomalies from excluded anomalies. (\Cref{fig:metrics}-a)}
For each class $c\in\mathcal C$, 
we use the definition $z_c=z_{\{c\}}$ that 
specifies only class $c$ as anomalous.
We evaluate the scores $s(z_c)$ using $\mathcal A_c$ as positives
and $\mathcal A_{\neg c}=\mathcal A\setminus\mathcal A_c$ as negatives.
Averaging the metric values equally over classes gives
\begin{equation}
  \label{eq:dc_disc}
  {\DCDisc}_m
  =
  \frac{1}{|\mathcal C|}
  \sum_{c\in\mathcal C}
  m\bigl(
    s(z_c);\mathcal A_c,\mathcal A_{\neg c}
  \bigr).
\end{equation}
Normal frames do not enter this comparison,
so strong separation from normal frames cannot mask poor discrimination
between target and excluded anomalies.

\paragraph{\bDCDetDelta{}: measure the matching-definition difference. (\Cref{fig:metrics}-b)}
For each target class $c\in\mathcal C$,
we use $\mathcal A_c$ as positives and $\mathcal N$ as negatives.
We evaluate this comparison under the matching definition $z_c$
and each non-matching definition $z_k$,
where $k\in\mathcal C\setminus\{c\}$.
We subtract the mean metric value under non-matching definitions
from the value under $z_c$, then average equally over target classes:
\begin{equation}
  \label{eq:dc_det_delta}
  {\DCDetDelta}_m
  =
  \frac{1}{|\mathcal C|}
  \sum_{c\in\mathcal C}
  \left[
    m\bigl(
      s(z_c);\mathcal A_c,\mathcal N
    \bigr)
    -
    \frac{1}{|\mathcal C|-1}
    \sum_{k\in\mathcal C\setminus\{c\}}
    m\bigl(
      s(z_k);\mathcal A_c,\mathcal N
    \bigr)
  \right].
\end{equation}
A detector satisfying \Cref{eq:definition_blind_scores}
receives zero, regardless of how well it separates
anomalous frames from normal frames.

\paragraph{\bDCSelDelta{}: select the matching event in a multi-event video. (\Cref{fig:dissociation})}
We measure whether the matching definition improves separation
of the target event from other anomalous events in the same video.
For each multi-event video $V_n$,
let $\mathcal C_n\subseteq\mathcal C$ be the anomaly classes present,
with $|\mathcal C_n|\ge2$.
Let $\mathcal A_{n,c}$ contain the frame indices $i=(n,\tau)$
labeled as class $c$ in $V_n$.
The other anomalous frames in that video form
\begin{equation}
  \label{eq:dc_sel_negatives}
  \mathcal A_{n,\neg c}
  =
  \bigcup_{k\in\mathcal C_n\setminus\{c\}}
  \mathcal A_{n,k}.
\end{equation}
We use $\mathcal A_{n,c}$ as positives
and $\mathcal A_{n,\neg c}$ as negatives under both the matching
definition $z_c$ and each non-matching definition $z_k$,
where $k\in\mathcal C_n\setminus\{c\}$.
Let $\Omega_{\mathrm{sel}}$ contain one pair $(n,c)$
for each evaluated video $V_n$ and each target $c\in\mathcal C_n$.
We average the matching-definition difference equally over these pairs:
\begin{equation}
  \label{eq:dc-sel-delta}
  \begin{aligned}
    {\DCSelDelta}_m
    &=
    \frac{1}{|\Omega_{\mathrm{sel}}|}
    \sum_{(n,c)\in\Omega_{\mathrm{sel}}}
    \Biggl[
      m\bigl(
        s(z_c);\mathcal A_{n,c},\mathcal A_{n,\neg c}
      \bigr) -
      \frac{1}{|\mathcal C_n|-1}
      \sum_{k\in\mathcal C_n\setminus\{c\}}
      m\bigl(
        s(z_k);\mathcal A_{n,c},\mathcal A_{n,\neg c}
      \bigr)
    \Biggr].
  \end{aligned}
\end{equation}
Normal frames may remain in the video as context
but are excluded from the metric.
A detector satisfying \Cref{eq:definition_blind_scores}
receives zero even if it accurately localizes all anomalous events.

\begin{figure*}[t]
  \centering
  \includegraphics[width=\textwidth]{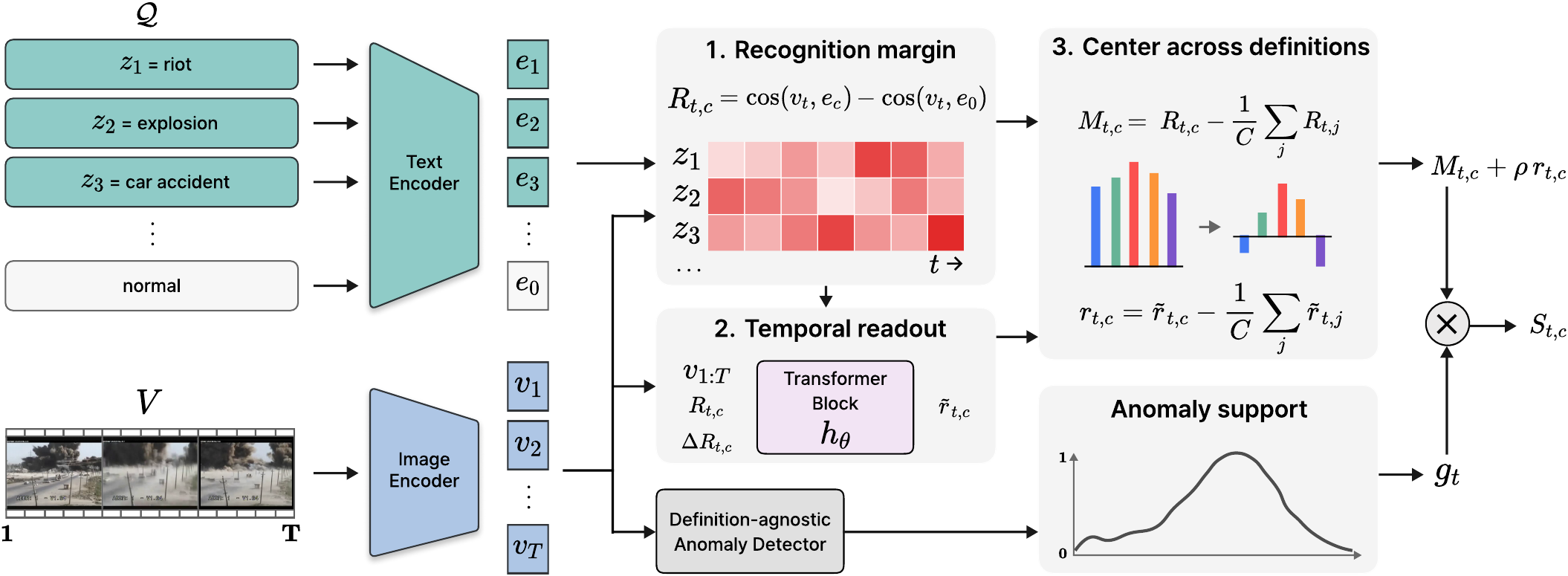}
  \caption{
    DeCoS converts anomaly support into definition-contrastive scores.
    A frozen definition-agnostic detector estimates anomaly support $g_t$. 
    The readout produces a zero-sum contrast $M_{t,c}+\rho r_{t,c}$ across the definitions in $\mathcal Z$. 
    Their product preserves anomaly support while preventing a definition-shared score offset.
  }
  \label{fig:architecture}
\end{figure*}

\section{DeCoS: Definition-Contrastive Scoring}
\label{sec:method}
DeCoS is a definition-relative scorer.
It evaluates a queried definition against a comparison set $\mathcal Z$.
This set-conditioned interface is also used by existing language-based
VAD models, which compare video features against multiple textual labels~\citep{wu2024vadclip,lagovad}.

\paragraph{Scoring.}
We score the supplied definition $z_c$ within a comparison set
$\mathcal Z=\{z_1,\ldots,z_C\}$, where $C$ is the number of comparison entries.
For a video $V$, let $v_t$ denote its CLIP visual feature
at feature time step $t$.
Let $e_c$ and $e_0$ be the text embeddings
of $z_c$ and the normal reference, respectively.
We compute recognition margins relative to normality
and remove their mean across definitions:
\begin{align}
  R_{t,c}
  &=
  \cos(v_t,e_c)
  -
  \cos(v_t,e_0),
  \label{eq:recognition_margin_rewrite}\\
  M_{t,c}
  &=
  R_{t,c}
  -
  \frac{1}{C}\sum_{k=1}^{C}R_{t,k}.
  \label{eq:centered_margin_rewrite}
\end{align}

A temporal readout $h_\theta$, shared across definitions,
uses the visual sequence, recognition margins,
and their first-order temporal differences
to produce bounded residuals $\tilde r_{t,c}\in[-1,1]$.
Let $g_t\in[0,1]$ be the frozen detector's anomaly support
aligned to the feature sequence and shared across definitions.
We center the residuals and combine the refined evidence with this support:
\begin{align}
  r_{t,c}
  &=
  \tilde r_{t,c}
  -
  \frac{1}{C}\sum_{k=1}^{C}\tilde r_{t,k},
  \label{eq:temporal_residual_centered_rewrite}\\
  S_{t,c}
  &=
  g_t\left(M_{t,c}+\rho r_{t,c}\right),
  \label{eq:DeCoS_score_rewrite}
\end{align}
where $\rho>0$ is a fixed residual scale.
Centering both evidence terms gives
$\sum_{c=1}^{C}S_{t,c}=0$,
so the readout cannot raise all definition scores together.

\paragraph{Training.}
We train the readout on single-event and multi-event clips
using PreVAD video-level labels~\citep{lagovad}.
At each supervised time step,
cross-entropy over the comparison definitions uses
$\alpha(M_{t,1:C}+\rho r_{t,1:C})$ as logits
and the matching definition as its target.
We optimize $\theta$ and the logit scale $\alpha$,
averaging first over time steps within each supervised region and then over regions in the batch.
Original clips inherit their video-level label at all valid time steps;
in synthesized clips, only the inserted event regions enter the loss.
The readout must distinguish matching from non-matching definitions
rather than merely detect anomalous events.
The readout architecture, training construction, full loss,
and frame alignment are detailed in Appendix~\ref{app:DeCoS_details}.

\section{Experimental Results}
\label{sec:experimental_results}
We now test whether the distinction between anomaly detection and definition following
changes how existing VAD systems should be evaluated.
We first examine how Drift@5 combines target-versus-normal
and target-versus-other-anomaly discrimination.
We then evaluate definition following directly across model families,
test definition-conditioned selection on natural multi-event videos,
and examine whether the results depend on exact event names.
Finally, we ablate DeCoS to connect its components to the behaviors they are designed to support.

\paragraph{Protocol.}
We evaluate on UCF-Crime (UCF), XD-Violence (XD),
and MSAD~\citep{sultani2018,wu2020not,zhu2024advancing}.
The comparison includes weakly supervised VAD, open-vocabulary VAD,
language-guided OWVAD, and generative vision--language models~\citep{llava,zhu2025internvl3,bai2025qwen25,bai2025qwen3}.
For a fair zero-shot comparison,
VadCLIP, LaGoVAD, PLOVAD, and DeCoS use the same PreVAD training source~\citep{wu2024vadclip,lagovad,plovad}.
Model-specific inputs and scoring rules are given in Appendix~\ref{app:baseline_details}.
We evaluate all three metrics with both AUROC and AP.
The main tables report AUROC for compactness, with the corresponding AP results in Appendix~\ref{app:full_metric_tables}.

\begin{table*}[t]
\centering
\small
\caption{
\textbf{Drift@5 can obscure differences in anomaly discrimination.}
\DfiveNormal{} and \DfiveExcluded{} average
$\mathrm{AUROC}^{\mathrm{normal}}(\Gamma)$ and
$\mathrm{AUROC}^{\mathrm{excluded}}(\Gamma)$, respectively,
over the five definitions used for \Dfive{}.
}
\label{tab:drift5_decomposition}
\resizebox{\textwidth}{!}{%
\begin{tabular}{l|ccc|ccc|ccc}
\toprule
&
\multicolumn{3}{c|}{$\Dfive_{\mathrm{AUROC}}$ $(\uparrow)$}
&
\multicolumn{3}{c|}{$\DfiveNormal_{\mathrm{AUROC}}$ $(\uparrow)$}
&
\multicolumn{3}{c}{$\DfiveExcluded_{\mathrm{AUROC}}$ $(\uparrow)$}
\\
Method
& UCF & XD & MSAD
& UCF & XD & MSAD
& UCF & XD & MSAD
\\
\midrule

LaGoVAD
& \textbf{79.7} & 85.9 & \underline{85.2}
& \textbf{80.8} & 90.3 & \underline{89.9}
& 49.9 & 55.4 & 49.4 \\

VadCLIP
& 78.2 & \underline{89.2} & \textbf{86.8}
& 78.9 & \textbf{91.6} & \textbf{90.7}
& \underline{58.9} & \underline{73.4} & \underline{57.5} \\

\textbf{DeCoS}
& \underline{78.5\(\pm\)3.7}
& \textbf{90.5\(\pm\)1.3}
& 85.2\(\pm\)0.9
& \underline{79.1\(\pm\)3.8}
& \underline{90.9\(\pm\)1.2}
& 88.3\(\pm\)0.9
& \textbf{64.6\(\pm\)0.6}
& \textbf{88.1\(\pm\)1.9}
& \textbf{62.3\(\pm\)1.3} \\

\bottomrule
\end{tabular}
}
\end{table*}

\begin{table*}[t]
\centering
\scriptsize
\caption{
\textbf{Definition-conditioned evaluations separate generic anomaly detection from definition following.}
\DCDisc{} values are AUROC percentages; $\Delta$ values are AUROC point margins.
}
\label{tab:def_following_eval}
\setlength{\tabcolsep}{3pt}
\renewcommand{\arraystretch}{0.85}
\resizebox{\textwidth}{!}{%
\begin{tabular}{@{}l|ccc|ccc|ccc@{}}
\toprule
&
\multicolumn{3}{c|}{$\DCDisc_{\mathrm{AUROC}}$ $(\uparrow)$}
&
\multicolumn{3}{c|}{$\DCDetDelta_{\mathrm{AUROC}}$ $(\uparrow)$}
&
\multicolumn{3}{c}{$\DCSelDelta_{\mathrm{AUROC}}$ $(\uparrow)$}
\\
Method
& UCF & XD & MSAD
& UCF & XD & MSAD
& UCF & XD & MSAD
\\
\midrule

VadCLIP
& 52.0 & 60.8 & 59.4
& 3.2 & 5.0 & 2.3
& 8.7 & 21.9 & 14.2 \\

LaGoVAD
& 45.2 & 41.0 & 49.0
& 0.0 & -0.1 & 0.0
& -0.2 & -0.7 & 0.0 \\

PLOVAD
& 47.8 & 52.3 & 48.0
& 0.0 & 0.0 & 0.0
& 0.0 & 0.0 & 0.0
\\[-0.2ex]

\midrule

LLaVA-1.5 7B
& 56.4 & 55.4 & 59.9
& 6.6 & 6.1 & 9.8
& 6.0 & 8.9 & 20.1
\\

InternVL3 8B
& 64.9 & \underline{71.1} & 71.0
& 14.5 & \underline{20.9} & \underline{18.8}
& 26.7 & \underline{36.2} & 35.3
\\

Qwen2.5-VL 7B
& 62.5 & 68.4 & 68.5
& 13.2 & 19.2 & 17.5
& 21.8 & 34.2 & 36.1
\\

Qwen3-VL 8B
& 61.3 & 66.5 & 67.4
& 11.7 & 17.7 & 16.7
& 20.0 & 27.1 & 31.4
\\[-0.2ex]

\midrule

\textbf{DeCoS}
& \textbf{76.1\(\pm\)0.8}
& \textbf{87.1\(\pm\)2.1}
& \textbf{78.3\(\pm\)1.8}
& \textbf{30.0\(\pm\)0.1}
& \textbf{49.2\(\pm\)1.7}
& \textbf{37.1\(\pm\)2.8}
& \textbf{39.1\(\pm\)2.1}
& \textbf{53.5\(\pm\)3.6}
& \textbf{49.6\(\pm\)1.7}
\\

\bottomrule
\end{tabular}
}
\end{table*}

\subsection{High Drift@5 does not imply definition following}
\label{sec:exp_drift5_failure}

\Cref{tab:drift5_decomposition} shows that large improvements
in separating target anomalies from excluded anomalies
can produce much smaller gains in Drift@5.
On XD-Violence, DeCoS exceeds LaGoVAD by $32.7$ AUROC points
on \DfiveExcluded{}, but by only $4.6$ points on \Dfive{}.
The difference on \DfiveNormal{} is just $0.6$ points.
On MSAD, DeCoS exceeds LaGoVAD by $12.8$ points on \DfiveExcluded{},
yet both models obtain $85.2\%$ on \Dfive{}.
Thus, higher Drift@5 does not necessarily indicate
better discrimination between target and excluded anomalies.

\subsection{Definition-conditioned metrics reveal definition blindness}
\label{sec:exp_definition_metrics}

\Cref{tab:def_following_eval} reveals large differences in definition following
that are not apparent from Drift@5 alone.
On XD-Violence, VadCLIP obtains $89.2\%$ AUROC under Drift@5,
but only $60.8\%$ on \DCDisc{},
with \DCDetDelta{} and \DCSelDelta{} values of $5.0$ and $21.9$ points.
LaGoVAD and PLOVAD show little or no matching-definition gain across the three benchmarks.
In contrast, InternVL3 and Qwen2.5-VL show substantially larger gains, 
indicating that models with strong anomaly-detection performance
can differ sharply in how their scores respond to the supplied definition.

DeCoS achieves the highest values on all three AUROC-based metrics
across UCF-Crime, XD-Violence, and MSAD.
Relative to the best baseline in each comparison,
it improves \DCDisc{} by $7.3$--$16.0$ AUROC points,
\DCDetDelta{} by $15.5$--$28.3$ points,
and \DCSelDelta{} by $12.4$--$17.3$ points.

\begin{table}[t]
  \centering
  \scriptsize
  \caption{
  \textbf{Definition-conditioned performance persists beyond event-name.}
  Left: evaluated event names are absent from the training vocabulary.
  Right: event names are removed and replaced by natural-language definitions.
  Each name-free entry averages the metrics from three descriptions.
}
  \label{tab:novel_namefree}
  \setlength{\tabcolsep}{2.5pt}
  \renewcommand{\arraystretch}{1}
  \begin{minipage}[t]{0.49\linewidth}
    \centering
    \resizebox{\linewidth}{!}{%
      \begin{tabular}{@{}l|ccc|ccc@{}}
        \toprule
        \multicolumn{7}{c}{(a) \textbf{Unseen event names}} \\
        \midrule
        &
        \multicolumn{3}{c|}{$\DCDisc_{\mathrm{AUROC}}$}
        &
        \multicolumn{3}{c}{$\DCDetDelta_{\mathrm{AUROC}}$}
        \\
        Method
        & UCF & XD & MSAD
        & UCF & XD & MSAD
        \\
        \midrule

        LaGoVAD
        & 44.3 & 15.2 & 45.4 & 0.0 & -0.2 & 0.0 \\

        VadCLIP
        & \underline{48.1} & \underline{28.7} & \underline{54.0} & \underline{2.2} & \underline{2.3} & \underline{1.5} \\

        \textbf{DeCoS}
        & \textbf{75.8} & \textbf{75.6} & \textbf{75.4} & \textbf{29.2} & \textbf{29.9} & \textbf{31.7} \\

        \bottomrule
      \end{tabular}
    }
  \end{minipage}
  \hfill
  \begin{minipage}[t]{0.49\linewidth}
    \centering
    \resizebox{\linewidth}{!}{%
      \begin{tabular}{@{}l|ccc|ccc@{}}
        \toprule
        \multicolumn{7}{c}{(b) \textbf{Name-free definitions}} \\
        \midrule
        &
        \multicolumn{3}{c|}{$\DCDisc_{\mathrm{AUROC}}$}
        &
        \multicolumn{3}{c}{$\DCDetDelta_{\mathrm{AUROC}}$}
        \\
        Method
        & UCF & XD & MSAD
        & UCF & XD & MSAD
        \\
        \midrule

        LaGoVAD
        & 45.2
        & 40.7
        & 49.0
        & -0.1
        & -0.2
        & +0.0
        \\

        VadCLIP
        & \underline{50.6}
        & \underline{52.3}
        & \underline{55.4}
        & \underline{1.6}
        & \underline{3.0}
        & \underline{0.9}
        \\

        \textbf{DeCoS}
        & \textbf{72.0} & \textbf{86.3} & \textbf{71.8} & \textbf{27.4} & \textbf{53.0} & \textbf{28.3} \\

        \bottomrule
      \end{tabular}
    }
  \end{minipage}
  \vspace{-1.0em}
\end{table}

\subsection{Definition-conditioned selection in natural multi-event videos}
\label{sec:exp_selection}

\begin{wraptable}{r}{0.30\linewidth}
  \centering
  \small
  \vspace{-1.3em}
      \caption{
Natural multi-event evaluation in AUROC points.
  }
  \label{tab:real_xd_dc_sel}
  \begin{tabular}{@{}lc@{}}
    \toprule
    Method & \DCSelDelta{} \\
    \midrule
    VadCLIP & \underline{19.2} \\
    LaGoVAD & 1.2 \\
    PLOVAD & 0.0 \\
    LLaVA-1.5 7B & 0.6 \\
    InternVL3 8B & -2.6 \\
    Qwen2.5-VL 7B & -3.2 \\
    Qwen3-VL 8B & -6.1 \\
    \midrule
    \textbf{DeCoS}
    & $\mathbf{36.1{\pm}8.6}$ \\
    \bottomrule
  \end{tabular}
  \vspace{-3.3em}
\end{wraptable}

The controlled \DCSelDelta{} evaluation uses synthesized two-event clips.
We next ask whether the differences in definition following observed in this controlled setting
also appear in natural multi-event videos.
We apply the same \DCSelDelta{} comparison to $43$ XD-Violence videos
containing at least two anomaly classes in separable intervals,
yielding $89$ target-label queries.
DeCoS obtains $36.1$ AUROC points, compared with $19.2$ for VadCLIP.
All other baselines obtain at most $1.2$ points,
while the generative vision--language models produce near-zero or negative margins
(\Cref{tab:real_xd_dc_sel}).
These results show that the differences in definition following observed in the controlled evaluation
persist in natural multi-event videos.

\subsection{Definition following beyond fixed event names}
\label{sec:exp_semantic_controls}

A user-specified definition need not use an event name seen during training.
We therefore evaluate two settings that break this correspondence.
In the unseen-name setting, the evaluated event names do not appear in the PreVAD training labels (\Cref{tab:novel_namefree}-a).
In the name-free setting, the event name is removed entirely and replaced by natural-language visual descriptions (\Cref{tab:novel_namefree}-b).
DeCoS retains strong definition-conditioned performance in both settings, 
while LaGoVAD remains near zero on the $\Delta$ metrics and VadCLIP shows much smaller gains.
These results show that DeCoS's definition following is not limited to exact event-name matching.

\subsection{Ablation study}
\label{sec:exp_constructive_validation}

\begin{wraptable}{r}{0.48\linewidth}
  \vspace{-2.6em}
    \caption{
Component ablation on XD-Violence.
}
  \label{tab:DeCoS_ablation_xd_wrap}
  \centering
  \scriptsize
  \setlength{\tabcolsep}{2.3pt}
  \renewcommand{\arraystretch}{0.86}

  \resizebox{\linewidth}{!}{%
    \begin{tabular}{@{}l|ccc@{}}
      \toprule
      Variant
      & $\DCDisc$
      & $\DCDetDelta$
      & $\DCSelDelta$
      \\
      \midrule

      DeCoS
      & ${87.1{\pm}2.1}$ & ${49.2{\pm}1.7}$ & ${53.5{\pm}3.6}$ \\

      $-$residual
      & $76.7{\pm}0.0$ & $40.9{\pm}0.0$ & $48.8{\pm}0.0$ \\

      $-$centering
      & $84.1{\pm}2.3$ & $45.5{\pm}0.8$ & $50.3{\pm}1.2$ \\

      $-$zero-sum
      & $82.6{\pm}5.0$ & $40.5{\pm}1.9$ & $52.7{\pm}4.6$ \\

      $-$multi-event
      & $80.4{\pm}2.7$ & $44.4{\pm}4.2$ & $37.3{\pm}3.6$ \\
      
      \bottomrule
    \end{tabular}
  }
  \vspace{-3.0em}
\end{wraptable}
Table~\ref{tab:DeCoS_ablation_xd_wrap} ablates the components of DeCoS on XD-Violence.
The $-$centering and $-$zero-sum variants remove the mean subtraction from the base margin and residual, respectively.
Each variant uses a separately trained checkpoint with the same definition-agnostic anomaly detector.
Removing the residual readout gives the largest drop in \DCDisc{}.
Removing residual centering reduces \DCDetDelta{} by $8.7$ points.
Removing base centering reduces all three metrics on XD-Violence.
Removing multi-event training examples reduces \DCSelDelta{} by $16.2$ points.


\section{Discussion and Limitations}
\label{sec:discussion}

\paragraph{Definition following should be evaluated separately.}
Changing the anomaly definition during evaluation is not, by itself, enough to evaluate definition following.
In Drift@5 and C-VAD-style evaluation, target anomalies are compared against a negative pool containing both normal frames and anomalies excluded by the current definition.
Because normal frames make up most of this pool on the benchmarks we study, a detector can score well by separating anomalies from normality even if its ranking changes little with the definition.
OWVAD evaluation should therefore report definition following separately from conventional anomaly detection.

\paragraph{Definition sensitivity is not definition following.}
A model can respond to changes in the input text without ranking the event specified by that text more highly.
Definition following therefore requires a directional comparison: on the same visual evidence, the matching definition should rank the matching event higher than non-matching definitions do.
This is why our $\Delta$ metrics compare matching and non-matching definitions
rather than measuring score variation alone.


\paragraph{The evaluated definitions remain benchmark-grounded.}
We evaluate definition following across the anomaly concepts in UCF-Crime, XD-Violence, and MSAD, including settings where the exact event name is unseen during training or removed from the test-time definition. These controls show that definition following need not depend on exposure to the exact event name during training or its inclusion in the test-time definition. They do not establish generalization to arbitrary user definitions or entirely new anomaly concepts, since some events with unseen names are conceptually close to those represented in PreVAD and our evaluation remains anchored to benchmark events. Evaluating definition following with more open-ended user definitions is an important direction for future benchmarks.

\paragraph{Natural multi-event evaluation remains limited in scale.}
Our controlled two-event clips provide a clean test of whether changing the definition changes the relative scores of co-occurring anomalous events.
We also observe the same distinction on $43$ naturally multi-event XD-Violence videos, but this subset is small and less balanced than the controlled evaluation.
A larger benchmark with naturally co-occurring anomalies and query-specific temporal annotations would allow definition-conditioned selection to be evaluated in natural videos at a larger scale.

\section{Conclusion}
\label{sec:conclusion}

Open-world video anomaly detection should be evaluated by whether temporal scores follow the supplied anomaly definition, 
not only by whether anomalous frames are separated from normal ones.
We show that existing temporal evaluations can obscure this distinction.
Across UCF-Crime, XD-Violence, and MSAD, normal frames make up $87.9\%$--$96.4\%$ of the negative pool under Drift@5 and $78.6\%$--$93.0\%$ under C-VAD evaluation.
As a result, strong target-versus-normal separation can contribute substantially to the reported score even when a model poorly distinguishes the target anomaly from other anomalies.
Our definition-conditioned evaluation confirms this distinction across multiple model families: strong generic anomaly detection can coexist with weak definition following.

These findings change what should count as evidence of OWVAD behavior.
Definition following should be evaluated explicitly alongside conventional anomaly detection, rather than inferred from text conditioning or strong target-versus-rest performance.
DeCoS provides a constructive example that definition following can be improved with definition-relative scoring.
More generally, testing how anomaly scores change when the anomaly definition changes provides a direct way to determine whether an OWVAD system follows the supplied anomaly definition.

\subsection*{AI use statement}
Generative AI tools were used to assist with sentence editing, grammar checking, typographical checks, isolated utility functions, 
and generation of candidate descriptions used in the name-free evaluation.
The authors reviewed every AI-assisted change, 
verified the mathematical definitions and all reported values against the evaluation outputs, and 
take responsibility for the final text, claims, code, and artifacts.
No reported experimental result was generated or selected by a language model.

\subsection*{Ethics statement}
This work uses existing public video anomaly detection datasets and pretrained models and does not collect new human-subject data.
Video anomaly detection may be used in surveillance settings, where deployment can raise privacy and fairness concerns.
Our experiments are limited to benchmark evaluation and do not study or make claims about deployment decisions involving individuals.

\subsection*{Reproducibility statement}
The main paper provides the definitions of all proposed evaluation metrics and the experimental protocols.
The appendix specifies the DeCoS architecture, training and inference procedure, baseline scoring rules, implementation and runtime settings, AP counterparts of the main results, and the definitions used in the semantic controls.
These details are intended to support reproduction of both the evaluation and the reported experiments.


\bibliography{iclr2027_conference}
\bibliographystyle{iclr2027_conference}

\clearpage
\appendix
\section{Additional Evaluation Details}
\label{app:evaluation_details}

\paragraph{Drift@5.}
XD-Violence and MSAD use the five class subsets in LaGoVAD's Table D.
UCF-Crime uses a local extension: five balanced seven-class subsets of its
13 anomaly classes, constructed with seed $7$.
For each method, the subset score is the maximum over the selected
definition-conditioned score columns, using the same scoring rule as in
the main evaluation. Anomaly spans inherit the video's single
class label; the five subset metrics are averaged equally.

\paragraph{Controlled selection.}
For each ordered pair of distinct classes, we sample $20$ pairs of test
anomaly clips with seed $7$. At least five annotated feature steps are
required; extracted segments are capped at $32$ steps.
Each feature sequence is constructed as $[N_0,A,N_1,B,N_2]$, with normal
segments sampled at lengths of $24$--$44$ steps (or the entire normal clip
when shorter). Each sequence is evaluated with both events as the target;
this swaps the target and competitor masks, not the temporal sequence.
The resulting $3120/600/2200$ sequences for UCF/XD/MSAD yield
$6240/1200/4400$ directed evaluations, averaged equally.
For DeCoS, the query occupies slot $0$ and is replaced between the matching and
competing evaluations. Four filler definitions, sampled from classes other
than the two events, remain fixed. Normal context is excluded from the metric.

\paragraph{Natural selection.}
The XD-Violence evaluation uses label-specific intervals from the released
multi-class annotations and contains $43$ videos and $89$ target-label
queries ($40$ videos with two queries and three with three).
Intervals are mapped to the feature grid by flooring the start and rounding
the end upward at the $16$-frame interval.
For each target, positives are its annotated steps excluding overlap with
any competing label; negatives are the union of competing-label steps
excluding the target. The matching-definition metric minus the mean
competing-definition metric is averaged equally over target-label queries.
DeCoS uses the full XD anomaly vocabulary in this evaluation.

\subsection{AP under Target-vs-Rest Evaluation}
\label{app:ap_decomposition}

The main text notes that AP uses the same mixed negative pool as
target-vs-rest AUROC, but does not admit the same additive decomposition.
Here we give the exact expression.

For a target frame $p\in\mathcal A_\Gamma$, let
$T_\Gamma(p)$ denote the number of target frames with scores at least
$s_p(z_\Gamma)$, and let $B_\Gamma^{N}(p)$ and $B_\Gamma^{E}(p)$ denote
the corresponding numbers of normal frames and excluded-anomaly frames.
Then
\begin{equation}
\mathrm{AP}^{\mathrm{rest}}(\Gamma)
=
\frac{1}{|\mathcal A_\Gamma|}
\sum_{p\in\mathcal A_\Gamma}
\frac{T_\Gamma(p)}
{T_\Gamma(p)+B_\Gamma^{N}(p)+B_\Gamma^{E}(p)}.
\label{eq:app_d5_ap}
\end{equation}

Normal frames and excluded anomalies therefore enter the same precision
denominator, even though their contributions cannot be separated into
the weighted average used for AUROC.

\section{DeCoS Implementation Details}
\label{app:DeCoS_details}

\subsection{Comparison Set and Text Features}
\label{app:DeCoS_inputs}

DeCoS evaluates the queried anomaly definition within a comparison set
$\mathcal Z=\{z_1,\ldots,z_C\}$, where $C$ is the number of comparison entries.
For the full-benchmark evaluation, $\mathcal Z$ contains the anomaly definitions
of the corresponding dataset, without an additional normal column.
Controlled selection instead uses the query and four fixed fillers.
When only one anomaly definition is supplied, we additionally include a normal
definition as a second comparison entry ($C=2$), so that the queried definition
is compared against normality.

We use frozen CLIP ViT-B/16 image and text encoders.
Let $\mathbf v_t\in\mathbb R^{512}$ denote the normalized visual feature at
feature time step $t$, and let $\mathbf e_c$ denote the normalized text embedding
of definition $z_c$.
Each anomaly definition is encoded using the template
\[
  \text{``a video of \{\}''}.
\]
The normal reference $\mathbf e_0$ is the normalized mean embedding of three
fixed descriptions of normal video content.
The single-query comparison entry uses the text ``a video of normal''.
The recognition margin used in the main text is
\begin{equation}
  R_{t,c}
  =
  \cos(\mathbf v_t,\mathbf e_c)
  -
  \cos(\mathbf v_t,\mathbf e_0).
  \label{eq:app_recognition_margin}
\end{equation}




\subsection{Residual Readout}
\label{app:readout}
We obtain the anomaly support $g_t$ from the frozen pre-fusion,
definition-agnostic binary head of LaGoVAD.
This branch receives only CLIP visual features and no anomaly definition.
We apply its sigmoid output to obtain $g_t\in[0,1]$.

The temporal readout uses the CLIP visual sequence together with the
recognition margin and its temporal derivative.
We first project each $512$-dimensional visual feature to $256$ dimensions
and process the sequence with a two-layer Transformer encoder:
\begin{equation}
  \mathbf H_{1:T}
  =
  E_{\mathrm{vis}}(W_v\mathbf v_{1:T}),
  \qquad
  W_v:\mathbb R^{512}\rightarrow\mathbb R^{256}.
  \label{eq:app_visual_encoder}
\end{equation}
Both Transformer encoders used by DeCoS have hidden dimension $256$,
four attention heads, feed-forward dimension $512$, and dropout $0.1$.

For each definition $c$, the readout also uses the recognition margin
$R_{t,c}$ and its temporal derivative
\begin{equation}
\Delta R_{t,c}
=
\begin{cases}
R_{2,c}-R_{1,c}, & t=1,\\
(R_{t+1,c}-R_{t-1,c})/2, & 1<t<T,\\
R_{T,c}-R_{T-1,c}, & t=T.
\end{cases}
\label{eq:app_margin_derivative}
\end{equation}
These two scalars are concatenated with $\mathbf H_t$ and projected back
to $256$ dimensions:
\begin{equation}
  \mathbf u_{t,c}
  =
  \operatorname{GELU}
  \left(
  W_f[\mathbf H_t;R_{t,c};\Delta R_{t,c}]
  +\mathbf b_f
  \right).
  \label{eq:app_residual_input}
\end{equation}

A second two-layer Transformer encoder, with parameters shared across
definitions, processes each definition-specific temporal sequence:
\begin{align}
  \mathbf h^{\mathrm{res}}_{1:T,c}
  &=
  E_{\mathrm{res}}(\mathbf u_{1:T,c}),\\
  \tilde r_{t,c}
  &=
  \tanh
  \left(
  \mathbf w_r^\top\mathbf h^{\mathrm{res}}_{t,c}+b_r
  \right).
  \label{eq:app_raw_residual}
\end{align}
The residual head is initialized with zero weights and bias.
There is no attention across definitions: the same
$E_{\mathrm{res}}$ is applied independently to each definition sequence.

We center both the recognition margin and the residual
over the comparison set:
\begin{align}
  M_{t,c}
  &=
  R_{t,c}
  -
  \frac{1}{C}\sum_{k=1}^{C}R_{t,k},\\
  r_{t,c}
  &=
  \tilde r_{t,c}
  -
  \frac{1}{C}\sum_{k=1}^{C}\tilde r_{t,k}.
\end{align}

Let $g_t\in[0,1]$ denote the anomaly support produced by the frozen
definition-agnostic detector and aligned to the CLIP feature sequence.
The final definition-conditioned score is
\begin{equation}
  S_{t,c}
  =
  g_t\left(M_{t,c}+\rho r_{t,c}\right),
  \qquad
  \rho=0.45.
  \label{eq:app_final_score}
\end{equation}

\subsection{Training and Inference}
\label{app:training_inference}

We train only the definition-conditioned readout using PreVAD video-level annotations.
The CLIP encoders and the definition-agnostic detector remain frozen, and no component is adapted on UCF-Crime, XD-Violence, or MSAD.
Training examples include original single-event clips, one-event splices with normal context, and two-event splices containing distinct anomaly categories.
For original anomaly clips, all valid feature time steps inherit the clip-level anomaly definition.
For synthesized clips, the known insertion locations determine the supervised time steps and their matching definitions.
Each example is paired with $C=6$ anomaly definitions, including the matching definition and distractor definitions.
Normal frames are retained as temporal context.
Synthetic normal context is excluded from the selection loss; original
clips retain weak video-level supervision even when some of their steps
depict normal activity. No additional human frame-level labels are used.

Let $\mathcal R$ be the supervised regions in a batch: all valid steps of
an original anomaly clip, or each inserted event region of a synthetic clip.
For region $r$ in example $q(r)$, let $y_r\in\{1,\ldots,C\}$ be its
video-level event label. Training uses ungated readout logits with learned
scale $\alpha$:
\begin{equation}
  L_{t,c}^{(q)}
  =
  \alpha\left(M_{t,c}^{(q)}+\rho r_{t,c}^{(q)}\right).
  \label{eq:app_training_logits}
\end{equation}
The readout is trained with cross-entropy over the $C$ anomaly-definition
columns:
\begin{equation}
  \mathcal L_{\mathrm{sel}}
  =
  \frac{1}{|\mathcal R|}
  \sum_{r\in\mathcal R}
  \frac{1}{|r|}\sum_{t\in r}
  \operatorname{CE}
  \left(
    L_{t,1:C}^{(q(r))},
    y_r
  \right).
  \label{eq:app_selection_loss}
\end{equation}
The logit scale $\alpha$ is initialized to $10$ and optimized jointly with
the readout.

At inference time, full-vocabulary evaluations retain the same comparison
set; controlled selection retains the fillers while replacing the query.
The visual encoder and the definition-agnostic detector are evaluated once per video, while the margin and residual branches are applied to all definitions using shared parameters.
Each definition is processed independently along the temporal dimension, with
no attention between definitions.
For a queried definition $z_c$, DeCoS returns the corresponding temporal score
sequence $S_{1:T,c}$.

The detector support $g_t$ and the DeCoS readout are evaluated on the same
CLIP feature grid.
To obtain frame-level scores, we repeat each feature-level score over the
corresponding video frames.
Let $\pi_d(\tau)\in\{1,\ldots,T\}$ map frame $\tau$ in dataset $d$ to its
corresponding feature block.
Before smoothing, this gives
\begin{equation}
  S_\tau(V,z_c)
  =
  S_{\pi_d(\tau),c}.
  \label{eq:app_frame_alignment}
\end{equation}
The feature interval is $8$ frames for UCF-Crime and MSAD and $16$ frames
for XD-Violence.
For all DeCoS evaluations, Gaussian smoothing is
applied to the gated score after frame repetition, averaging standard
deviations of $1$--$9$ feature intervals.

\subsection{Implementation and Efficiency}
\label{app:implementation_summary}
\label{app:efficiency}

The DeCoS readout contains $2.3$M trainable parameters.
Together with the $6.5$M frozen detector, the deployed scorer contains
$8.8$M parameters, excluding CLIP.
We optimize it with AdamW using a learning rate of $2\times10^{-4}$,
weight decay of $10^{-4}$, and gradient clipping at $1.0$.
The residual scale is fixed to $\rho=0.45$, and the learned logit scale
$\alpha$ is initialized to $10$.
The main benchmark tables report mean and standard deviation over three training seeds.

The evaluation server has four NVIDIA RTX 6000 Ada GPUs ($48$\,GB each)
and two AMD EPYC 9124 CPUs.

\begin{table}[t]
  \centering
  \small
  \setlength{\tabcolsep}{4pt}
  \caption{
    Inference efficiency of feature-based methods on the median-length anomalous MSAD video.
    Parameter counts and GFLOPs exclude the shared frozen CLIP backbone.
    Timings average $20$ repetitions at batch size one; DeCoS processes
    all dataset definitions in one pass and excludes temporal smoothing.
    End-to-end runtime is the post-CLIP time plus the shared $23.8$\,ms
    feature-extraction measurement.
  }
  \label{tab:efficiency_feature}
  \begin{tabular}{lcccc}
    \toprule
    Method
    & Scorer params
    & After CLIP (ms)
    & GFLOPs
    & E2E (ms) \\
    \midrule
    VadCLIP & 9.5M & 83 & 176 & 107 \\
    LaGoVAD & 19.1M & 98 & 73 & 122 \\
    PLOVAD & 4.0M & 10 & 66 & 34 \\
    DeCoS & 8.8M & 10 & 6.7 & 34 \\
    \bottomrule
  \end{tabular}
\end{table}

\begin{table}[t]
  \centering
  \small
  \setlength{\tabcolsep}{5pt}
  \caption{
    Inference cost of the generative VLM baselines on the same MSAD video.
    GFLOPs and runtime cover full-model inference.
  }
  \label{tab:efficiency_vlm}
  \begin{tabular}{lccc}
    \toprule
    Method & Model size & GFLOPs & E2E (ms) \\
    \midrule
    LLaVA-1.5 & 7B & 588{,}148 & 8{,}829 \\
    InternVL3 & 8B & 340{,}253 & 9{,}923 \\
    Qwen2.5-VL & 7B & 1{,}870{,}142 & 36{,}287 \\
    Qwen3-VL & 8B & 1{,}259{,}435 & 21{,}421 \\
    \bottomrule
  \end{tabular}
\end{table}

\subsection{Effect of Query-Class Count}
\label{app:single_query_normal}
\label{app:single_query_interface}
\label{app:query_class_count}

DeCoS is evaluated in the main paper with the full dataset vocabulary as the comparison set. At deployment, however, a user may provide only a small number of anomaly definitions. We therefore vary the number of queried anomaly classes $c$ at test time and keep the trained checkpoint fixed.

The only degenerate case is a single anomaly class without any competing column: centering would make the definition-relative score identically zero. For the operational one-query interface, we therefore include a normal column,
\[
  \mathcal Q=\{z_0,z_1\},\qquad z_0=\text{normal},
\]
so the single supplied anomaly still competes against normality. Before the residual correction,
\[
  M_{t,1}=\frac{1}{2}(R_{t,1}-R_{t,0}),\qquad
  M_{t,0}=-M_{t,1}.
\]
All rows with $c\ge2$ use anomaly-class comparison sets of size $c$, sampled as in the candidate-count sweep. We report DC-Disc and DC-Det$\Delta$ only; DC-Sel$\Delta$ uses the splice-specific comparison-set construction and is not the same cardinality axis.

\begin{table}[h]
  \centering
  \small
  \setlength{\tabcolsep}{4pt}
  \caption{Effect of the number of queried anomaly classes. The $c=1$ row adds a normal comparison column. Intermediate sizes average five candidate-set draws per seed; the one-query and full-vocabulary rows use one set. Values are means and standard deviations across three seed-level means, in AUROC percentages or AUROC-point margins.}
  \label{tab:app_query_class_count}
  \begin{tabular}{@{}llcc@{}}
    \toprule
    Dataset & Query classes $c$ & DC-Disc AUROC & DC-Det$\Delta$ AUROC \\
    \midrule
    \multirow{6}{*}{UCF}
      & $1$ + normal & $69.3{\pm}1.8$ & $22.0{\pm}1.3$ \\
      & 2 & $67.3{\pm}0.4$ & $21.1{\pm}0.7$ \\
      & 3 & $71.7{\pm}0.2$ & $26.6{\pm}0.7$ \\
      & 5 & $74.4{\pm}0.4$ & $29.3{\pm}0.9$ \\
      & 8 & $76.8{\pm}0.2$ & $31.2{\pm}0.5$ \\
      & 13 (full) & $76.1{\pm}0.8$ & $30.0{\pm}0.1$ \\
    \midrule
    \multirow{5}{*}{XD}
      & $1$ + normal & $79.7{\pm}0.3$ & $32.3{\pm}0.7$ \\
      & 2 & $80.2{\pm}1.2$ & $39.3{\pm}1.2$ \\
      & 3 & $84.2{\pm}1.4$ & $45.3{\pm}1.1$ \\
      & 5 & $87.1{\pm}1.9$ & $48.6{\pm}1.6$ \\
      & 6 (full) & $87.1{\pm}2.1$ & $49.2{\pm}1.7$ \\
    \midrule
    \multirow{6}{*}{MSAD}
      & $1$ + normal & $72.5{\pm}2.7$ & $26.2{\pm}0.9$ \\
      & 2 & $70.8{\pm}1.5$ & $27.3{\pm}1.7$ \\
      & 3 & $73.7{\pm}1.0$ & $31.1{\pm}1.6$ \\
      & 5 & $76.5{\pm}1.2$ & $34.3{\pm}1.9$ \\
      & 8 & $77.5{\pm}1.9$ & $36.2{\pm}2.9$ \\
      & 11 (full) & $78.3{\pm}1.8$ & $37.1{\pm}2.8$ \\
    \bottomrule
  \end{tabular}
\end{table}

The trend is the expected one: more competing anomaly definitions usually sharpen definition following, while the smallest operational setting already remains strong. In particular, the $c=1$ row is above the trained VAD/OWVAD baselines in Table~\ref{tab:def_following_eval}, and its DC-Det$\Delta$ margin is larger than every non-DeCoS method there. Thus DeCoS does not require a large user-provided candidate set to improve definition-conditioned scoring.

\section{Baseline Scoring Details}
\label{app:baseline_details}

All baselines are evaluated with the same anomaly definitions used in the
definition-conditioned evaluation.
For each definition $z_c$, we obtain a temporal score sequence
$S_{1:T,c}$ as follows.

\paragraph{PLOVAD.}
We use the official PLOVAD architecture and losses, retrained on PreVAD.
Its temporal anomaly score $a_t$ does not depend on the supplied definition.
We therefore use
\begin{equation}
  S_{t,c}=a_t
  \qquad
  \text{for all } c.
  \label{eq:app_plovad_score}
\end{equation}

\paragraph{VadCLIP.}
We use the official VadCLIP architecture and losses, retrained on PreVAD.
For definition-conditioned evaluation, we use the frame-level alignment
scores produced by its A-branch as \(S_{t,c}\).

\paragraph{LaGoVAD.}
We use the LaGoVAD model trained on PreVAD released by author~\cite{lagovad}.
Its binary detection head produces a single temporal anomaly logit for the
entire supplied anomaly-definition set, rather than a separate score for each definition.
It therefore does not provide a per-definition anomaly score when multiple
anomaly definitions are supplied.
For each queried definition, we evaluate the model separately with
$[\texttt{normal}, z_c]$ and use the sigmoid of the binary anomaly logit as
$S_{t,c}$.
Prompt sampling is disabled during evaluation.

\paragraph{Generative vision--language models.}
We evaluate LLaVA-1.5, InternVL3, Qwen2.5-VL, and Qwen3-VL as frozen
frame-level scorers.
For each sampled frame and definition $z_c$, 
we use the same numeric-rating
instruction:
\begin{quote}
\small
\texttt{Is the following happening in this video: \{z\}? Reply with ONLY a
single number between 0.00 and 1.00 (e.g. 0.80).}
\end{quote}
The first generated number in $[0,1]$ is used as $S_{t,c}$.
Unparseable responses receive score $0$.
Model-specific chat templates wrap the same instruction.

\section{AP Counterparts to Main Results}
\label{app:full_metric_tables}

The main tables report AUROC for compactness.
Here we report the corresponding AP results using the same evaluation
protocols and score sequences.

\begin{table*}[t]
  \centering
  \small
  \setlength{\tabcolsep}{3.2pt}
  \caption{
    AP counterpart of Table~\ref{tab:drift5_decomposition}.
    Values are AP percentages.
    DeCoS results are mean and standard deviation over three seeds.
  }
  \label{tab:app_drift5_ap}
  \resizebox{\textwidth}{!}{
    \begin{tabular}{l|ccc|ccc|ccc}
      \toprule
      &
      \multicolumn{3}{c|}{$\Dfive_{\mathrm{AP}}$ $(\uparrow)$}
      &
      \multicolumn{3}{c|}{$\DfiveNormal_{\mathrm{AP}}$ $(\uparrow)$}
      &
      \multicolumn{3}{c}{$\DfiveExcluded_{\mathrm{AP}}$ $(\uparrow)$}
      \\
      Method
      & UCF & XD & MSAD
      & UCF & XD & MSAD
      & UCF & XD & MSAD
      \\
      \midrule
      LaGoVAD
      & 12.8 & 41.9 & 37.3 & 14.5 & 57.7 & 52.3 & 53.8 & 56.9 & 56.5 \\
      VadCLIP
      & 15.1 & 55.1 & 44.7 & 16.3 & 59.6 & 56.3 & 65.1 & 77.6 & 65.4 \\
      DeCoS
      & $\mathbf{19.8{\pm}2.8}$ & $\mathbf{64.4{\pm}1.5}$ & $\mathbf{45.1{\pm}1.6}$ & $\mathbf{20.8{\pm}3.1}$ & $\mathbf{67.9{\pm}1.3}$ & $\mathbf{56.7{\pm}1.1}$ & $\mathbf{69.3{\pm}0.6}$ & $\mathbf{84.5{\pm}1.1}$ & $\mathbf{66.2{\pm}1.3}$ \\
      \bottomrule
    \end{tabular}
  }
\end{table*}

\begin{table*}[t]
  \centering
  \small
  \setlength{\tabcolsep}{4pt}
  \caption{
    AP counterpart of the \DCDisc{} and \DCDetDelta{} results in
    Table~\ref{tab:def_following_eval}.
    \DCDisc{} values are AP percentages, and \DCDetDelta{} values are
    AP-point margins.
  }
  \label{tab:app_def_following_ap}
  \resizebox{\textwidth}{!}{%
    \begin{tabular}{l|ccc|ccc}
      \toprule
      &
      \multicolumn{3}{c|}{$\DCDisc_{\mathrm{AP}}$ $(\uparrow)$}
      &
      \multicolumn{3}{c}{$\DCDetDelta_{\mathrm{AP}}$ $(\uparrow)$}
      \\
      Method
      & UCF & XD & MSAD
      & UCF & XD & MSAD
      \\
      \midrule
      VadCLIP
      & 13.5 & 44.8 & 20.2 & 2.6 & 16.6 & 8.6 \\
      LaGoVAD
      & 8.6 & 19.2 & 11.4 & 0.0 & 0.0 & 0.0 \\
      PLOVAD
      & 10.3 & 21.2 & 11.6
      & 0.0 & 0.0 & 0.0
      \\
      \midrule
      LLaVA-1.5 7B
      & 10.2 & 18.9 & 14.5
      & 0.9 & 3.0 & 8.6
      \\
      InternVL3 8B
      & 20.7 & 36.4 & 29.7
      & 8.7 & 17.6 & \textbf{29.8}
      \\
      Qwen2.5-VL 7B
      & 15.2 & 31.1 & 24.4
      & 2.4 & 14.6 & 22.7
      \\
      Qwen3-VL 8B
      & 15.1 & 31.5 & 25.7
      & 2.9 & 14.4 & 22.6
      \\
      \midrule
      DeCoS
      & $\mathbf{27.2{\pm}1.8}$ & $\mathbf{55.7{\pm}1.1}$ & $\mathbf{33.3{\pm}2.4}$ & $\mathbf{10.1{\pm}0.8}$ & $\mathbf{36.5{\pm}0.6}$ & $25.2{\pm}2.5$ \\
      \bottomrule
    \end{tabular}
  }
\end{table*}

\begin{table*}[t]
  \centering
  \small
  \setlength{\tabcolsep}{4pt}
  \caption{
    AP counterpart of the \DCSelDelta{} results in
    Table~\ref{tab:def_following_eval} and the natural-video evaluation in
    Table~\ref{tab:real_xd_dc_sel}.
    Values are AP-point margins.
  }
  \label{tab:app_dc_sel_ap}
  \begin{tabular}{l|ccc|c}
    \toprule
    &
    \multicolumn{3}{c|}{$\DCSelDelta_{\mathrm{AP}}$ $(\uparrow)$}
    &
    Natural XD $(\uparrow)$
    \\
    Method & UCF & XD & MSAD & XD \\
    \midrule
    VadCLIP
    & 6.0 & 14.5 & 9.8 & 13.5 \\
    LaGoVAD
    & -0.1 & -0.3 & 0.0 & 0.9 \\
    PLOVAD
    & 0.0 & 0.0 & 0.0 & 0.0
    \\
    \midrule
    LLaVA-1.5 7B
    & 2.7 & 3.9 & 9.7 & -0.4 \\
    InternVL3 8B
    & 12.8 & 16.3 & 16.6 & -1.2 \\
    Qwen2.5-VL 7B
    & 10.4 & 15.5 & 17.3 & -2.4 \\
    Qwen3-VL 8B
    & 9.3 & 11.9 & 14.3 & -2.7 \\
    \midrule
    DeCoS
    & $\mathbf{26.5{\pm}1.4}$ & $\mathbf{36.0{\pm}2.4}$ & $\mathbf{33.7{\pm}1.2}$ & $\mathbf{23.1{\pm}4.6}$ \\
    \bottomrule
  \end{tabular}
\end{table*}

\section{Semantic-Control Details}
\label{app:semantic_controls}

\subsection{Unseen Event Names}
\label{app:unseen_names}

We define an event name as unseen when the corresponding benchmark event name
does not appear in the PreVAD training labels.
\Cref{tab:app_unseen_names} lists the event names used in this evaluation.

\begin{table}[t]
  \centering
  \small
  \setlength{\tabcolsep}{4pt}
  \caption{
    Event names used in the unseen-name evaluation.
    The count for each dataset gives the number of unseen event names over the
    total number of anomaly categories.
  }
  \label{tab:app_unseen_names}
  \begin{tabular}{@{}p{0.18\linewidth}p{0.74\linewidth}@{}}
    \toprule
    Dataset & Unseen event names \\
    \midrule
    UCF-Crime $(9/13)$
    & Abuse, Arrest, Arson, Burglary, Fighting, RoadAccidents, Shooting,
      Shoplifting, Stealing \\
    XD-Violence $(3/6)$
    & Abuse, Fighting, Shooting \\
    MSAD $(6/11)$
    & Fighting, Object\_falling, People\_falling, Shooting,
      Traffic\_accident, Water\_incident \\
    \bottomrule
  \end{tabular}
\end{table}

\subsection{Name-Free Definitions}
\label{app:namefree_generation}

For the name-free evaluation, we replace each benchmark event name with three
short natural-language visual definitions.
The definition bank combines the project's existing visual definitions
with positive descriptions from GPT-4o-generated discriminative anchors.
The three selected descriptions per class are fixed for evaluation.

The benchmark event name is used only to generate these definitions and is not
included in the text provided to the evaluated model.
We evaluate each of the three definition sets independently and average
the resulting metrics, rather than averaging temporal scores before evaluation.
Scoring and temporal processing are unchanged from the class-name evaluation.
The complete definitions are listed in
Appendix~\ref{app:namefree_definitions}.

\subsection{Complete Name-Free Definitions}
\label{app:namefree_definitions}
The class names below are shown only as identifiers and are not included in
the text provided to the evaluated models.
The three definition sets are evaluated independently and their metrics
are averaged.
\begingroup
\small
\paragraph{UCF-Crime.}
\begin{description}[leftmargin=!,labelwidth=3.0cm,style=nextline,itemsep=0.3ex,topsep=0.3ex]
  \item[Abuse]
  \begin{enumerate}[leftmargin=1.3em,itemsep=0.1ex,topsep=0.1ex]
    \item someone mistreating or harming another person
    \item A person is striking another person with visible force and aggression.
    \item An individual is restraining another person against their will with clear physical dominance.
  \end{enumerate}
  \item[Arrest]
  \begin{enumerate}[leftmargin=1.3em,itemsep=0.1ex,topsep=0.1ex]
    \item police officers detaining and restraining a person
    \item Police officers place handcuffs on an individual's wrists while they are standing against a wall.
    \item A person is escorted by officers with hands restrained behind their back.
  \end{enumerate}
  \item[Arson]
  \begin{enumerate}[leftmargin=1.3em,itemsep=0.1ex,topsep=0.1ex]
    \item a person deliberately setting property on fire
    \item A person is seen igniting a flammable liquid on a structure with a lighter.
    \item An individual throws a lit object into a building, causing flames to erupt.
  \end{enumerate}
  \item[Assault]
  \begin{enumerate}[leftmargin=1.3em,itemsep=0.1ex,topsep=0.1ex]
    \item a violent physical attack on a person
    \item One person repeatedly strikes another with fists in a public area.
    \item A group surrounds an individual, pushing and hitting them aggressively.
  \end{enumerate}
  \item[Burglary]
  \begin{enumerate}[leftmargin=1.3em,itemsep=0.1ex,topsep=0.1ex]
    \item someone breaking into a building to take valuables
    \item A person is seen breaking a window to enter a building at night.
    \item An individual is carrying valuables out of a house through a forced-open door.
  \end{enumerate}
  \item[Explosion]
  \begin{enumerate}[leftmargin=1.3em,itemsep=0.1ex,topsep=0.1ex]
    \item a sudden violent burst of flames smoke and flying debris
    \item A sudden burst of flames and smoke rapidly expands outward from a central point.
    \item Debris and objects are violently propelled outward in all directions from an epicenter.
  \end{enumerate}
  \item[Fighting]
  \begin{enumerate}[leftmargin=1.3em,itemsep=0.1ex,topsep=0.1ex]
    \item people physically attacking each other
    \item Two individuals exchange rapid punches and kicks in close proximity.
    \item A person aggressively grabs another, pulling them forcefully towards the ground.
  \end{enumerate}
  \item[RoadAccidents]
  \begin{enumerate}[leftmargin=1.3em,itemsep=0.1ex,topsep=0.1ex]
    \item vehicles crashing or colliding on a road
    \item A vehicle collides with another vehicle, causing visible damage or sudden stop.
    \item A pedestrian is struck by a moving vehicle, resulting in immediate physical impact.
  \end{enumerate}
  \item[Robbery]
  \begin{enumerate}[leftmargin=1.3em,itemsep=0.1ex,topsep=0.1ex]
    \item someone taking belongings by force or threat
    \item A person forcefully grabs a bag from another individual and runs away.
    \item An individual points a weapon at a cashier while demanding money from the register.
  \end{enumerate}
  \item[Shooting]
  \begin{enumerate}[leftmargin=1.3em,itemsep=0.1ex,topsep=0.1ex]
    \item a person firing a gun
    \item A person is seen discharging a firearm with visible muzzle flash and recoil.
    \item An individual aims a gun and a projectile visibly impacts a target or surface.
  \end{enumerate}
  \item[Shoplifting]
  \begin{enumerate}[leftmargin=1.3em,itemsep=0.1ex,topsep=0.1ex]
    \item a customer secretly hiding store goods to avoid paying
    \item An individual conceals merchandise under clothing while glancing around for observers.
    \item A person quickly exits the store with items in hand, bypassing the checkout area.
  \end{enumerate}
  \item[Stealing]
  \begin{enumerate}[leftmargin=1.3em,itemsep=0.1ex,topsep=0.1ex]
    \item a person taking belongings that are not theirs
    \item A person discreetly places an item into their bag without paying.
    \item An individual quickly exits a store with merchandise in hand, bypassing the checkout.
  \end{enumerate}
  \item[Vandalism]
  \begin{enumerate}[leftmargin=1.3em,itemsep=0.1ex,topsep=0.1ex]
    \item a person deliberately damaging or defacing property
    \item A person is spray-painting graffiti on a public wall.
    \item An individual is smashing a car window with a blunt object.
  \end{enumerate}
\end{description}
\paragraph{XD-Violence.}
\begin{description}[leftmargin=!,labelwidth=3.0cm,style=nextline,itemsep=0.3ex,topsep=0.3ex]
  \item[Fighting]
  \begin{enumerate}[leftmargin=1.3em,itemsep=0.1ex,topsep=0.1ex]
    \item people physically attacking each other
    \item Two individuals are exchanging rapid punches and kicks in close proximity.
    \item One person is forcefully pushing another against a wall while raising a fist.
  \end{enumerate}
  \item[Shooting]
  \begin{enumerate}[leftmargin=1.3em,itemsep=0.1ex,topsep=0.1ex]
    \item a person firing a gun
    \item A person is seen aiming a firearm directly at another individual or group.
    \item A muzzle flash is visible as a firearm is discharged in a public setting.
  \end{enumerate}
  \item[Riot]
  \begin{enumerate}[leftmargin=1.3em,itemsep=0.1ex,topsep=0.1ex]
    \item a violent crowd disturbance
    \item Crowds of people are aggressively clashing with law enforcement, throwing objects and pushing barriers.
    \item Multiple individuals are seen vandalizing property and setting fires in a chaotic street scene.
  \end{enumerate}
  \item[Abuse]
  \begin{enumerate}[leftmargin=1.3em,itemsep=0.1ex,topsep=0.1ex]
    \item someone mistreating or harming another person
    \item A person is forcefully grabbing another's arm while the latter visibly struggles to break free.
    \item One individual is repeatedly striking another who is visibly trying to shield themselves.
  \end{enumerate}
  \item[Car accident]
  \begin{enumerate}[leftmargin=1.3em,itemsep=0.1ex,topsep=0.1ex]
    \item vehicles crashing or colliding on a road
    \item Two vehicles collide, causing visible damage and debris on the road.
    \item A car abruptly stops after hitting another vehicle, triggering airbags.
  \end{enumerate}
  \item[Explosion]
  \begin{enumerate}[leftmargin=1.3em,itemsep=0.1ex,topsep=0.1ex]
    \item a sudden violent burst of flames smoke and flying debris
    \item A sudden burst of flames and smoke rapidly expands outward from a central point.
    \item Debris is violently propelled in all directions following a bright flash.
  \end{enumerate}
\end{description}
\paragraph{MSAD.}
\begin{description}[leftmargin=!,labelwidth=3.0cm,style=nextline,itemsep=0.3ex,topsep=0.3ex]
  \item[Assault]
  \begin{enumerate}[leftmargin=1.3em,itemsep=0.1ex,topsep=0.1ex]
    \item a violent physical attack on a person
    \item One person repeatedly strikes another with fists in a public setting.
    \item A group surrounds an individual, pushing and hitting them aggressively.
  \end{enumerate}
  \item[Explosion]
  \begin{enumerate}[leftmargin=1.3em,itemsep=0.1ex,topsep=0.1ex]
    \item a sudden violent burst of flames smoke and flying debris
    \item A sudden burst of flames and smoke rapidly expands outward from a central point.
    \item Debris is violently ejected in all directions following a bright flash of light.
  \end{enumerate}
  \item[Fighting]
  \begin{enumerate}[leftmargin=1.3em,itemsep=0.1ex,topsep=0.1ex]
    \item people physically attacking each other
    \item Two individuals are exchanging rapid punches and kicks in close proximity.
    \item One person is aggressively grappling another, attempting to throw them to the ground.
  \end{enumerate}
  \item[Fire]
  \begin{enumerate}[leftmargin=1.3em,itemsep=0.1ex,topsep=0.1ex]
    \item flames and smoke spreading through a scene
    \item Flames visibly flicker and emit smoke from a concentrated area.
    \item Bright orange and red hues rapidly spread across a surface.
  \end{enumerate}
  \item[Object\_falling]
  \begin{enumerate}[leftmargin=1.3em,itemsep=0.1ex,topsep=0.1ex]
    \item an object dropping or toppling to the ground
    \item An object visibly detaches from a surface and descends rapidly to the ground.
    \item A stationary object suddenly moves downward without external force visible.
  \end{enumerate}
  \item[People\_falling]
  \begin{enumerate}[leftmargin=1.3em,itemsep=0.1ex,topsep=0.1ex]
    \item a person losing balance and falling down
    \item A person suddenly collapses to the ground without any external force.
    \item An individual loses balance and falls from a standing position to the floor.
  \end{enumerate}
  \item[Robbery]
  \begin{enumerate}[leftmargin=1.3em,itemsep=0.1ex,topsep=0.1ex]
    \item someone taking belongings by force or threat
    \item A person forcibly takes a bag from another individual who visibly resists.
    \item An individual brandishes a weapon while demanding items from a store clerk.
  \end{enumerate}
  \item[Shooting]
  \begin{enumerate}[leftmargin=1.3em,itemsep=0.1ex,topsep=0.1ex]
    \item a person firing a gun
    \item A person is seen aiming a firearm and discharging it with visible recoil.
    \item Muzzle flash is observed from a gun held by an individual.
  \end{enumerate}
  \item[Traffic\_accident]
  \begin{enumerate}[leftmargin=1.3em,itemsep=0.1ex,topsep=0.1ex]
    \item vehicles crashing or colliding on a road
    \item A vehicle collides with another, causing visible damage or abrupt stops.
    \item Pedestrians or cyclists are struck by a moving vehicle, resulting in sudden falls or injuries.
  \end{enumerate}
  \item[Vandalism]
  \begin{enumerate}[leftmargin=1.3em,itemsep=0.1ex,topsep=0.1ex]
    \item a person deliberately damaging or defacing property
    \item A person is spray-painting graffiti on a public wall in broad daylight.
    \item An individual is smashing car windows with a blunt object in a parking lot.
  \end{enumerate}
  \item[Water\_incident]
  \begin{enumerate}[leftmargin=1.3em,itemsep=0.1ex,topsep=0.1ex]
    \item a person struggling or submerged in water
    \item Water rapidly flooding an indoor space, visibly rising above floor level.
    \item A burst pipe spraying water forcefully across a room or corridor.
  \end{enumerate}
\end{description}
\endgroup

\end{document}